\documentclass[sn-mathphys-num]{sn-jnl}

\usepackage{graphicx}%
\usepackage{multirow}%
\usepackage{amsmath,amssymb,amsfonts}%
\usepackage{amsthm}%
\usepackage{mathrsfs}%
\usepackage[title]{appendix}%
\usepackage{xcolor}%
\usepackage{textcomp}%
\usepackage{manyfoot}%
\usepackage{booktabs}%
\usepackage{algorithm}%
\usepackage{algorithmicx}%
\usepackage{algpseudocode}%
\usepackage{listings}%

\theoremstyle{thmstyleone}%
\theoremstyle{thmstyletwo}%

\theoremstyle{thmstylethree}%

\begin{document}

\title[Article Title]{Holi-DETR: Holistic Fashion Item Detection Leveraging Contextual Information}


\author[1]{\fnm{Youngchae} \sur{Kwon}}\email{youngchaekwon@handong.ac.kr}

\author[1]{\fnm{Jinyoung} \sur{Choi}}\email{jychoi@handong.ac.kr}

\author*[2]{\fnm{Injung} \sur{Kim}}\email{ijkim@handong.edu}

\affil*[1]{\orgdiv{Department of CSEE}, \orgname{Handong Global University General Graduate School}, \orgaddress{\street{558 Handong-ro Buk-gu}, \city{Pohang}, \postcode{37554}, \state{Gyeongbuk}, \country{Republic of Korea}}}
\affil*[2]{\orgdiv{School of CSEE}, \orgname{Handong Global University}, \orgaddress{\street{558 Handong-ro Buk-gu}, \city{Pohang}, \postcode{37554}, \state{Gyeongbuk}, \country{Republic of Korea}}}


\abstract{
Fashion item detection is challenging due to the ambiguities introduced by the highly diverse appearances of fashion items and the similarities among item subcategories. To address this challenge, we propose a novel \textbf{Holi}stic \textbf{De}tection \textbf{Tr}ansformer (\textit{Holi-DETR}) that detects fashion items in outfit images holistically, by leveraging contextual information.
Fashion items often have meaningful relationships as they are combined to create specific styles.
Unlike conventional detectors that detect each item independently, Holi-DETR detects multiple items while reducing ambiguities by leveraging three distinct types of contextual information: (1) the co-occurrence relationship between fashion items, (2) the relative position and size based on inter-item spatial arrangements, and (3) the spatial relationships between items and human body key-points.
To this end, we propose a novel architecture that integrates these three types of heterogeneous contextual information into the Detection Transformer (DETR) and its subsequent models. In experiments, the proposed methods improved the performance of the vanilla DETR and the more recently developed Co-DETR by 3.6 percent points (pp) and 1.1 pp, respectively, in terms of average precision (AP).
}




\maketitle

\section{Introduction}\label{sec1}

Recent advances in deep learning technology have facilitated the widespread application of artificial intelligence (AI) technology in the fashion industry.
Fashion-related AI applications encompass various tasks, including fashion image classification \cite{seo2019hierarchical, tian2023improving}, fashion item detection \cite{mohammadi2021smart,ma2023efficient}, fashion item compatibility assessment \cite{cui2019dressing,sarkar2022outfittransformer}, fashion style classification, fashion item retrieval \cite{lin2020fashion,sarkar2022outfittransformer}, and virtual try-on \cite{han2018viton,islam2024deep}.
Among them, fashion item detection stands out for its significant industrial potential. It serves as the core components for a variety of applications, such as personalized recommendation, fashion trend analysis, and other functionalities that substantially enhance user experience.
While extensive research has been conducted on general object detection \cite{girshick2014rich,girshick2015fast,ren2016faster,liu2016ssd,carion2020end,zhu2020deformable,meng2021conditional,tian2022fully,li2022dn,liu2022dab,zhang2022dino,shehzadi20232d,hou2025relation}, relatively few studies have focused on detection methodologies specifically designed for fashion items \cite{lao2015convolutional,feng2018object,duan2019centernet,kim2021multiple,lee2021two,sidnev2021deepmark++,mohammadi2021smart,ma2023efficient}.

\begin{figure}[ht!]
\centering
\includegraphics[width=\columnwidth]{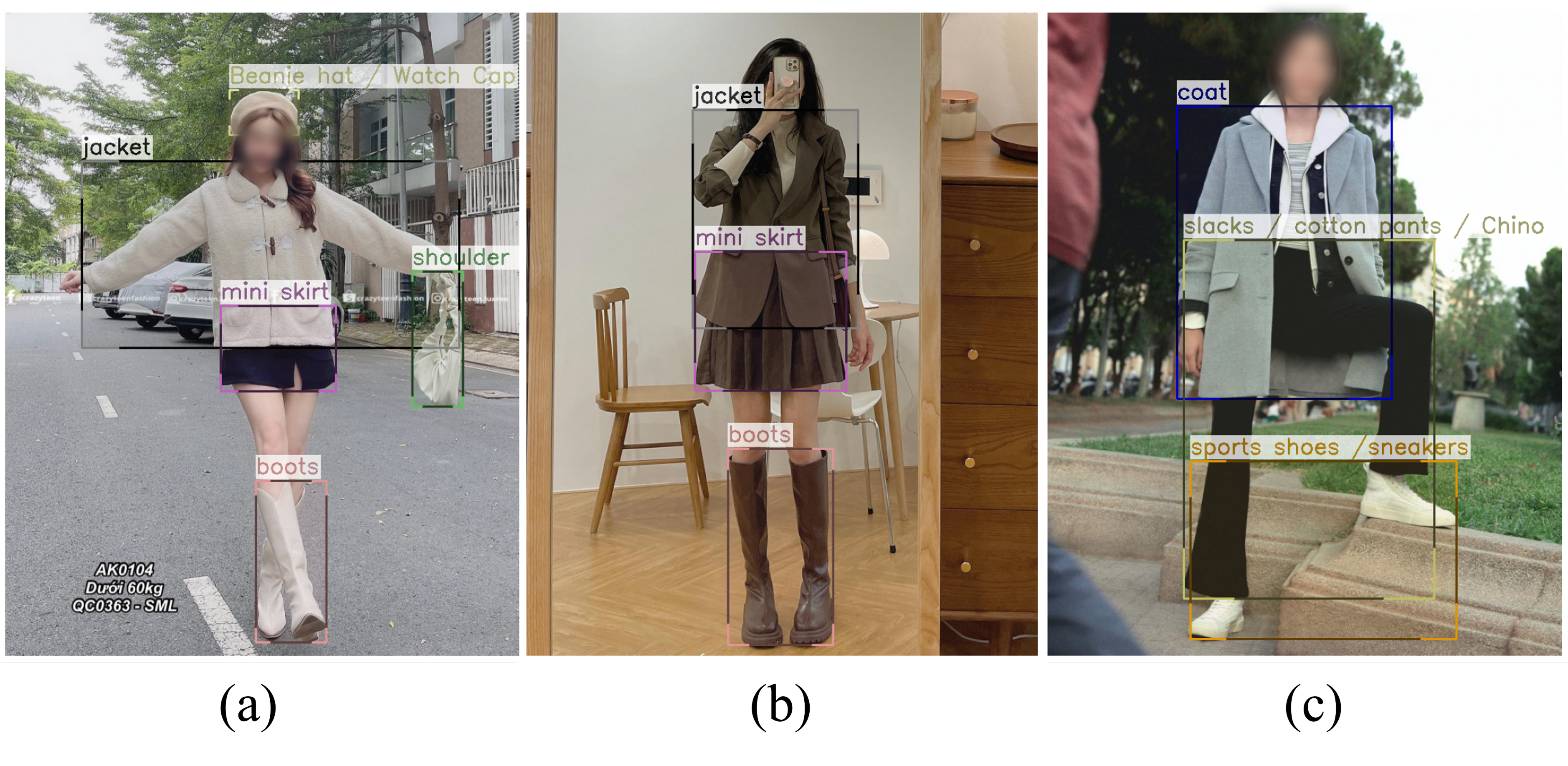}
\caption{Example of interclass similarity and intraclass diversity. The jacket in the (b) is more similar to the coat in the (c) than the jacket in the (a) in terms of collar, pockets, and shape of the garment}\label{fig1}
\end{figure}

Fashion item detection, encompassing item classification, presents a unique challenge due to the diverse shapes, materials, and colors of items, as well as the visual diversity introduced by different combination styles and wearer poses.
Furthermore, fashion item detection requires fine-grained classification. The top-level categories include outer, top, pants, skirt, dress, underwear, shoes, bag, cap/hat, accessory, and eyewear, each of which is further divided into more specific subcategories, as listed in Table \ref{tab1}.
Many subcategories share similar shapes and are separated by subtle differences. As depicted in Fig.~\ref{fig1}, visual diversity can be evident even within the same category, while similarities can also occur across different categories. The jackets in Fig.~\ref{fig1}(a) and (b) exhibit significant differences, yet the jacket in (b) shares visual similarity with the coat in (c). In addition, outfit images often include overlapping garments, such as tops layered with outerwear. The items inside are only partially visible because they are hidden by the items outside.
These characteristics of fashion images introduce significant ambiguities, rendering general object detection models ineffective for identifying fashion items and emphasizing the need for detection methodologies tailored to the fashion domain.


\begin{table}[ht!]
\caption{Fashion item categories in Showniq-H dataset}\label{tab1}
\begin{tabular}{c|l}
        \toprule
        Top-level categories & \multicolumn{1}{c}{Subcategories} \\
        \midrule
        \multirow{2}{*}{Outer} & Jacket, Coat, Jumper, Padding coat, Vest, Cardigan, Hood zip-up, \\ 
                                & Windbreaker \\
        \midrule
        \multirow{2}{*}{Top} & Casual-top, Sleeveless, Sweatshirt, Hooded T-shirt, Shirts/Blouse, \\
                                & Knit/Sweater, Tank top/Tube top, Crop top, Aodai, Polo shirt \\
        \midrule
        \multirow{2}{*}{Pants} & Shorts, Slacks/Cotton pants/Chino, Wide pants, Jeans, \\
                                & Cargo pants, Leggings, Jogger pants, Overalls, Jumpsuit \\
        \midrule
        {Skirt} & Mini skirt, Midi skirt, (Long) skirt \\
        \midrule
        {Dress} & Mini dress, Midi dress, (Long) dress \\
        \midrule
        {Swimwear} & Bikini, Rash guard, One-piece swimsuit, Beach tops, Beach pants \\
        \midrule
        \multirow{2}{*}{Underwear} & Bra, Panties/Underwear, Pajamas/Home wear, Slip dress, \\
                                & Bathrobe, Correction underwear \\
        \midrule
        \multirow{2}{*}{Shoes} & Shoes/Loafers, Heels, Sports shoes/Sneakers, Boots, \\
                                & Sandals/Slippers \\
        \midrule
        \multirow{3}{*}{Bag} & Crossbody bag/Messenger bag, Shoulder bag, Tote bag, \\
                                & Brief case, Waist bag, Canvas bag, Clutch bag, Backpack, \\
                                & Pouch, Wallet, Luggage \\
        \midrule
        \multirow{2}{*}{Cap/Hat} & Ballcap, Bucket hat, Fedora hat, Beanie hat/Watch cap, Sun cap, \\
                                & Hunting cap/Beret \\
        \midrule
        \multirow{2}{*}{Fashion accessory} & Phone case, Socks, Watch, Belt, Scarf, Tights/Stockings, Gloves, \\
                                & Mask, Hairband, Hairpin, Bandana, Hair tie \\
        \midrule
        {Eyewear} & Glasses, Sunglasses \\
        \bottomrule
    \end{tabular}
\end{table}

Many prior studies in fashion item detection rely on general object detection architectures \cite{lao2015convolutional}, such as Faster R-CNN \cite{ren2016faster} and its subsequent models \cite{alamsyah2019object,tian2023improving,li2024multi}. A few studies enhanced the region proposal network (RPN) \cite{hara2016fashion} to capture the hierarchy of the fashion item categories \cite{lee2021two}. However, existing models have limitations in dealing with the complexity and diversity of fashion items and outfit images.

We propose to address the aforementioned challenges through contextual object detection \cite{rabinovich2007objects,divvala2009empirical}, which incorporates various forms of contextual information into the detection process. 
In general, fashion items exhibit meaningful co-occurrence relationships because wearers often create specific styles by combining multiple fashion items. Moreover, certain item classes with similar appearance can be distinguished based on their relative position or size with respect to other items or human body key-points. For example, a midi skirt that appears similar to a long skirt can be effectively identified by referencing the distance between its bottom edge and the ankle. Therefore, referencing the context provided by other items or human body key-points can help reduce ambiguity and enhance discriminative capability in fashion item detection.

To this end, we present a novel \textbf{Holi}stic \textbf{De}tection \textbf{Tr}ansformer (\textit{Holi-DETR}). Holi-DETR extends the recently developed DETR \cite{carion2020end} by incorporating three types of contextual information extracted from fashion image data. One is the co-occurrence relationship between item classes that captures the likelihood of specific items appearing together. The other is the relative position and scale between items. The third is the positional relation between items and human body key-points.

Holi-DETR combines the three types of contextual information and uses them to adjust the attention weights of the self-attention sub-layers. Our method seamlessly integrates heterogeneous contextual information into the DETR architecture in a learnable manner. In addition, it is easily applicable to more recently developed DETR-based models, such as Co-DETR \cite{zong2023detrs}. To the best of our knowledge, Holi-DETR is the first contextual fashion item detector based on the Transformer architecture.


The main contributions of this study include the following:

\begin{enumerate}
\item The first holistic fashion item detector that incorporates three types of contextual information into a Transformer-based architecture.
\item A novel method that seamlessly integrates heterogeneous contextual information into the self-attention mechanism.
\item Improved fashion item detection performance on a realistic fashion image dataset, Showniq-H, that contains diverse fashion outfit images.
\end{enumerate}

This paper is structured as follows: Section \ref{sec2} introduces previous studies that are related to our work. Section \ref{sec3} describes the proposed methods for reflecting heterogeneous contextual information for fashion item detection. Section \ref{sec4} presents the results of quantitative and qualitative evaluation. Finally, Section \ref{sec5} concludes our work.

\section{Related Work}\label{sec2}

\subsection{Object Detection}\label{subsec1}

Deep learning-based object detection involves predicting the location and class of objects in an image or video, combining object classification and localization tasks. Most existing detection models are categorized into one of the two mainstream approaches: one-stage methods and two-stage methods. Two-stage detectors, including R-CNN \cite{girshick2014rich}, Fast R-CNN \cite{girshick2015fast}, and Faster R-CNN \cite{ren2016faster}, first generate candidate regions for objects and then refine their coordinates and classify them. While two-stage detectors generally exhibit higher accuracy, their detection speed is slower compared to one-stage detectors. In contrast, one-stage detectors, such as YOLO \cite{redmon2016you}, SSD \cite{liu2016ssd}, and FCOS \cite{tian2022fully}, perform localization and classification simultaneously, offering high speed and making them well-suited for real-time applications.

While the core concepts of these detection models were proposed long ago, their performance has been continuously improved through integration with powerful backbone networks, remaining widely used to this day. In particular, Transformer-based architectures such as ViT \cite{dosovitskiy2021an} and Swin Transformer \cite{liu2021swin} have been utilized as backbones for these detection models and significantly improved their performance. Nevertheless, these models have limitations that they rely on handcrafted components, such as anchors and the non-maximum suppression (NMS) algorithm, and that they are not trainable in an end-to-end manner.


Recently, Carion et al. \cite{carion2020end} proposed an end-to-end Detection Transformer (DETR), which eliminates the need for handcrafted components by addressing object detection as a direct set prediction problem. DETR first extracts embeddings from the input image using a Transformer encoder combined with a convolutional neural network (CNN) backbone. A Transformer decoder then generates output embeddings from a set of learnable object query vectors by leveraging the image embeddings through cross-attention. Finally, detection heads predict the coordinates and classes of the objects from the output embeddings. During training, DETR finds the minimum cost matching between the predicted and ground-truth objects through the Hungarian algorithm and computes the training loss based on the matching result.

DETR achieves a comparable performance to the conventional detection models but suffers from slow convergence. Numerous subsequent studies have been conducted to address this limitation and improve detection performance \cite{shehzadi20232d}. Deformable DETR \cite{zhu2020deformable} introduces a deformable attention mechanism inspired by deformable convolutions, while Conditional DETR  \cite{meng2021conditional} incorporates reference points to narrow the spatial range requiring attention for bounding box localization, thereby improving the efficiency of query learning. Liu et al. \cite{liu2022dab} replaced the object queries with dynamic anchor boxes, and Li et al. \cite{li2022dn} proposed a denoising training method to speed up the training of DETR integrated contrastive learning to DETR. Zhang et al. \cite{zhang2022dino} proposed an improved detection model that combines the key benefits of these prior studies while achieving greater efficiency and accuracy. Zong et al. \cite{zong2023detrs} applied a collaborative hybrid assignments training scheme that improves the training efficiency and detection performance.

\subsection{Contextual Object Detection}\label{subsec2}

Contextual object detection aims at improving the accuracy of multiple objects detection by leveraging contextual information derived from the objects themselves or the surrounding scene \cite{cinbis2012contextual}. Contextual object detectors utilizes various contextual information including the size and location of individual objects, the relationships between objects, and their interactions with the background \cite{divvala2009empirical,alamri2020improving}. Galleguillos and Belongie \cite{galleguillos2010context} categorized contextual information for object detection into three primary types: semantic, spatial, and scale.

Several previous studies have incorporated contextual information during the post-processing stage. These studies construct a graph in which the predicted objects are represented as nodes, and their relationships are depicted as edges. The predictions are then refined by incorporating contextual information through conditional random fields (CRF) \cite{rabinovich2007objects,galleguillos2008object,zolghadr2016scene}. They probabilistically estimate pairwise co-occurrence or spatial relationships and assign these as weights to the edges of the graph.

Recent studies have proposed end-to-end approaches that directly incorporate relationships between objects into the detection process.
Hu et al. \cite{hu2018relation} introduced the first end-to-end object detector that considers relationships between objects. They presented an object relation module capable of learning object relationships directly from data.
Barnea et al. \cite{barnea2019contextual} developed a probabilistic model that dynamically selects neighbors with high relevance to detection among surrounding objects. Alamri and Pugeault \cite{alamri2019contextual} proposed re-scoring and re-labeling techniques to generate a new embedding based on contextual information extracted from the detector's output. Pato et al. \cite{pato2020seeing} proposed a method to improve the performance of existing detectors by re-evaluating the detection confidence by utilizing context information of object detection using bidirectional recurrent neural network (RNN).

Few studies have been conducted to capture relationships between objects based on the Transformer architecture. Relation-enhanced DETR \cite{hao2023relation} incorporates a learnable matrix to model the relationships between object classes. Relation DETR \cite{hou2025relation} refines object coordinates in successive decoder layers by utilizing the relative positional information between the coordinates of corresponding objects predicted by the two preceding layers.

\subsection{Fashion Item Detection}\label{subsec2}

Object detection techniques have been applied in the fashion domain to identify fashion items and landmarks, such as key-points.
Most previous studies rely on well-established general object detection models and relatively simple CNN-based architectures \cite{feng2018object}. They involve fine-tuning pre-trained object detection models or incorporating specific modifications \cite{lao2015convolutional,kim2021multiple,lee2021two}. For instance, Sidnev et al. \cite{sidnev2021deepmark++} introduced a single-stage detector built on CenterNet \cite{duan2019centernet}. Several studies proposed detection methods specifically for fashion items \cite{mohammadi2021smart}, which involves predicting the attributes of clothing items in addition to their bounding boxes and categories \cite{lao2015convolutional}.

Nevertheless, most existing fashion item detectors fail to account for the unique characteristics of fashion image data outlined in Section \ref{sec1}. In particular, to the best of our knowledge, no prior study has addressed the ambiguities in fashion item detection by incorporating contextual information into Transformer-based models.

\section{Holistic Detection Transformer}\label{sec3}

\subsection{Overall Structure}\label{subsec3_1}

\begin{figure}[ht!]
\centering
\includegraphics[width=\columnwidth]{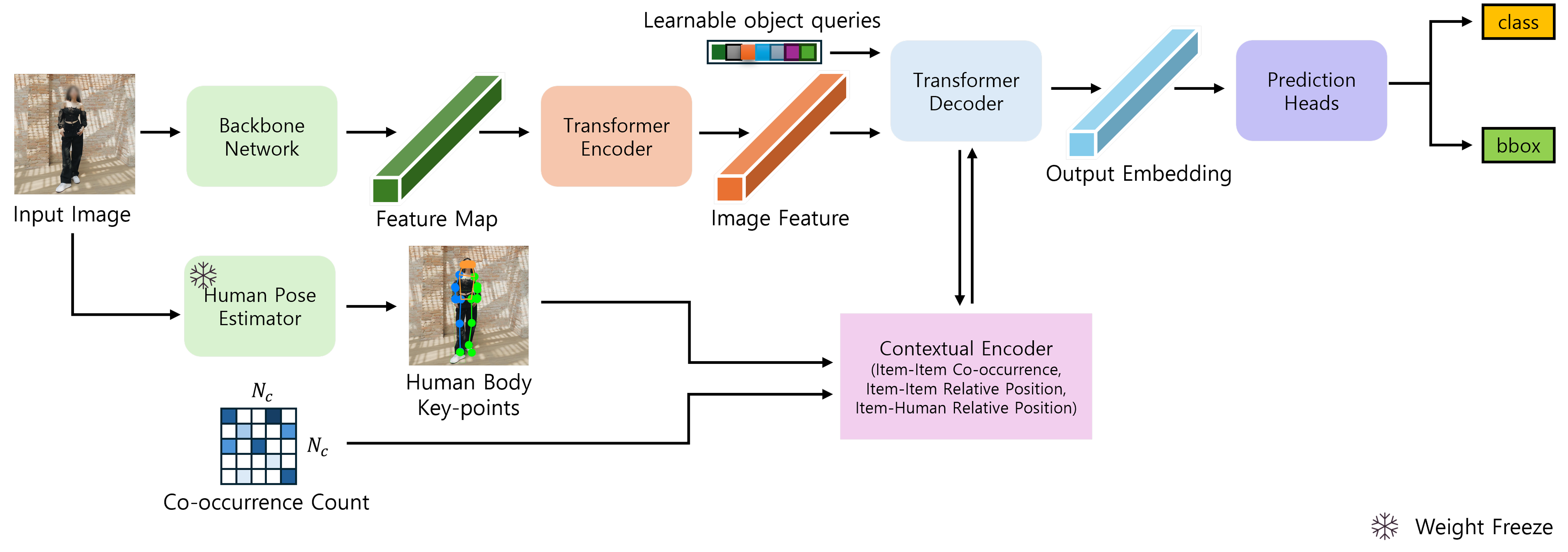}
\caption{The overall structure of Holi-DETR. The Transformer decoder generates output embeddings based on the object queries, the image features from the encoder, and three additional types of contextual information: item-item co-occurrence, item-item relational positions, and item-human relative positions}\label{fig2}
\end{figure}

The overall structure of the proposed model, Holi-DETR, is illustrated in Fig.~\ref{fig2}. Similar to DETR \cite{carion2020end}, Holi-DETR comprises a backbone network, a Transformer encoder, a Transformer decoder, and prediction heads. Additionally, it incorporates a human pose estimator and a context encoder to enhance self-attention weights with contextual information.

The backbone network and Transformer encoder extract image features. The Transformer decoder takes a set of object queries as input and generates an output embedding for each object query by leveraging the image features through cross attention. Additionally, each decoder layer includes prediction heads to classify objects and predict their bounding boxes. The final detection results are generated by the top layer, while the detection heads of other decoder layers assist training by providing auxiliary losses. The predictions of certain intermediate decoder layers are used to estimate contextual information.

The human pose estimator identifies key-points of the human body from the input image. In this study, we adopted a pre-trained human pose estimator \cite{xu2022vitpose} as described in Section \ref{subsec4_1}. The context encoder extracts the three types of contextual information based on the predictions of intermediate decoder layers and the estimated human body key-points. It then combines them, and encodes them into relation weights, which are added to the self-attention weights. To extract item-item co-occurrence matrices, the context encoder references a class-class co-occurrence count matrix. We computed the class-class co-occurrence count matrix from the training samples of the Showniq-H dataset introduced in Section \ref{sec4}.

\subsection{Context Encoder}

\begin{figure}[ht!]
\centering
\includegraphics[width=\columnwidth]{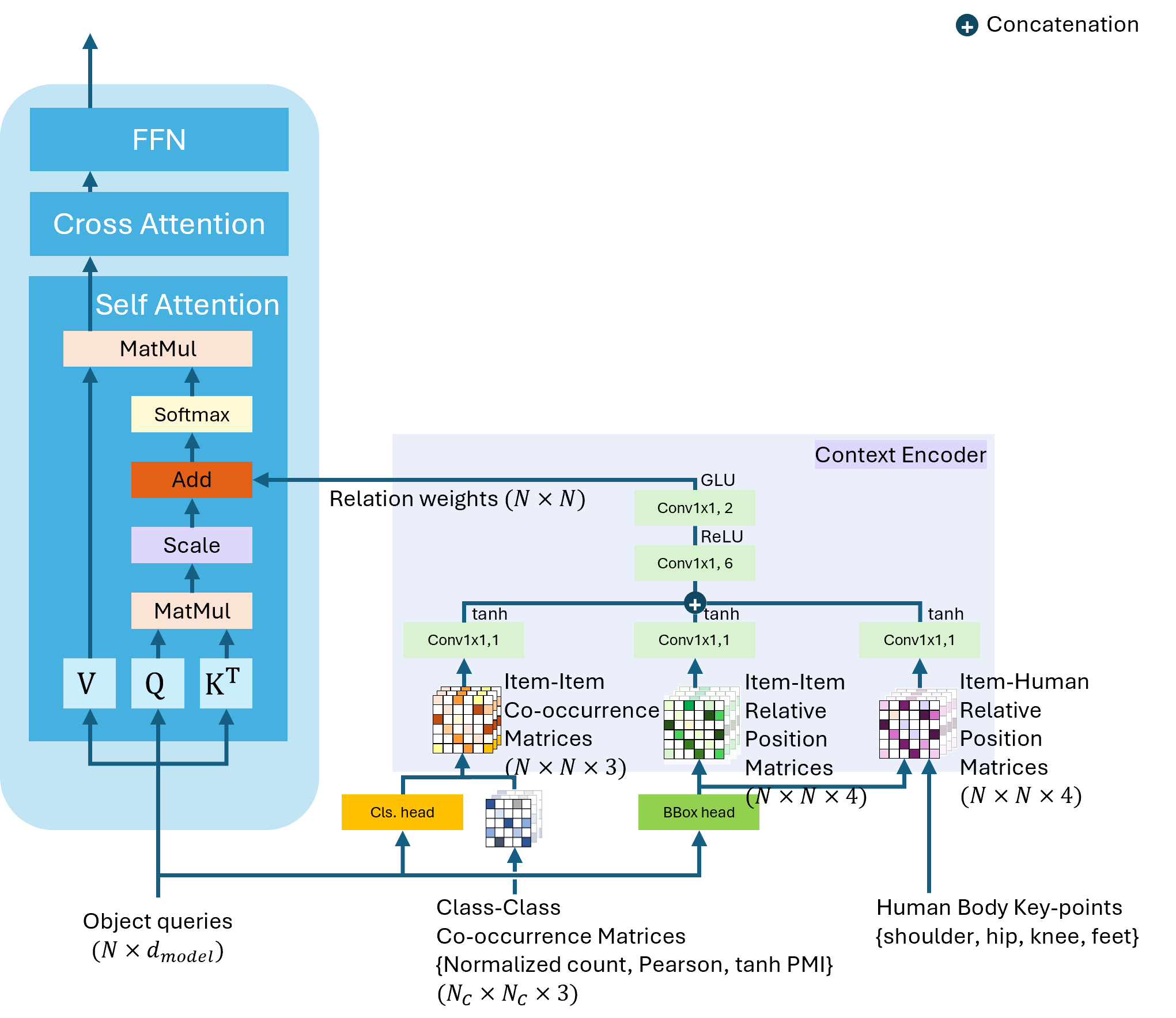}
\caption{Holi-DETR decoder layer. The context encoder extracts item-item co-occurrence matrices, item-item relative position matrices, and item-human relative position matrices, combines them, and encodes them into relation weights, which are then added to the self-attention weights. $N$ and $N_C$ denote the number of object queries and the number of classes, respectively, while $d_{model}$ refers to the feature dimension of the model }\label{fig3}
\end{figure}

As illustrated in Fig. \ref{fig3}, the context encoder computes relation weights and adds them to the self-attention weights of each decoder layer. To this end, the context encoder takes as input the prediction of the preceding layer, class-class co-occurrence matrices, and the estimated human body key-points. 
The context encoder retrieves item-item co-occurrence matrices from the class-class co-occurrence matrices based on the predicted classes from the object queries. It also computes item-item relative position matrices from the predicted bounding boxes and item-human relative position matrices using the predicted bounding boxes and the estimated human body key-points. The detailed procedures to compute the three types of context matrices are explained in the following subsections.

The context encoder then compresses the three types of context matrices, each consisting of multiple channels, into single-channel representations using $1\times1$ convolutions. It then sums the three matrices and encodes them into relation weights through two additional $1\times1$ convolutions. The relation weights are incorporated into the unnormalized attention weights prior to applying the softmax operation. The intermediate layers of the context encoder use the hyperbolic tangent (tanh) activation instead of ReLU to capture not only positive correlations but also negative correlations. For the final layer that computes the relation weights, we employed a gated linear unit (GLU), which serves as a gating mechanism to handle missing or incomplete information. 

While the accuracy of the relation weights depends on the predictions from preceding layers, the lower layers tend to produce less reliable predictions, which can potentially lead to training instability. To mitigate this issue, we incorporate contextual information only into the top three decoder layers.

\subsection{Co-occurrence Matrices}\label{subsec3_1}

We first compute the co-occurrence matrices of item classes to account for the co-occurrence relationships between items. The class-class co-occurrence count quantifies the frequency with which different classes co-occur.
Denoting the number of item classes as $N_c$, and the set of predicted object classes derived from the input image $I_k$ for $1 \leq k \leq M$ as $C_k \subseteq \{1,2,...,N_c\}$, the class-class co-occurrence count $A \in \mathbb{R}^{N_c \times N_c}$ is computed as Eq. \ref{eq1}. $u$ and $v$ are class indices.
\begin{align}
\label{eq1}
A[u,v] &= \sum_{k=1}^{M} A_k[u,v] \\
A_k[u,v] &= \textbf{1}[u \in C_k \text{ and } v \in C_k] \nonumber
\end{align}

Based on the co-occurrence count $A$, we compute three types of co-occurrence measurements between classes: normalized count, Pearson correlation coefficient, and tanh pointwise mutual information (PMI) \cite{church-hanks-1990-word}. The normalized co-occurrence count is computed as Eq. \ref{eq2} and represents the probability of co-occurrence for different item categories relative to each specific item category. Notably, this matrix is asymmetric.
\begin{equation}
R_{nc} = \frac{A[u,v]}{\sum_{v'=1}^{N_c} A[u,v']}
\label{eq2}
\end{equation}

The Pearson correlation coefficients represent the strength and direction of a linear relationship between two continuous variables. They are computed as Eq. \ref{eq3}, where $\sigma_u$ and $\sigma_v$ represent the standard deviation of the $u$-th and $v$-th elements.
\begin{equation}
R_{pcc} = \frac{Cov(A[:,u], A[:,v])}{\sigma_u \sigma_v}
\label{eq3}
\end{equation}

PMI quantifies the association between two events by comparing their joint probability with the product of their individual probabilities under the assumption of independence, as shown in Eq. \ref{eq4}. A higher PMI value indicates a stronger association between the events. The matrix captures both positive and negative correlations. Since the range of PMI values significantly differs from that of other correlation measurements, we apply the tanh function to constrain their range to [-1, +1], as shown in Eq. \ref{eq5}, where $N_t$ is the number of total occurrence and $f(u)$ and $f(v)$ represent the frequencies of the $u$-th and $v$-th elements, respectively. 

\begin{equation} 
R_{pmi} = \log{\frac{\text{Pr}(u,v)}{\text{Pr}(u)\text{Pr}(v)}}=\log{\frac{A[u,v] \cdot N_t}{f(u) \cdot f(v)}}
\label{eq4} 
\end{equation}

\begin{equation} 
R_{\tanh(pmi)} = \tanh{\left(R_{pmi}\right)}
\label{eq5} 
\end{equation}
Since the three co-occurrence measurements described above complement one another, we utilize all of them.

With the class IDs $(c_1,c_2,...,c_N)$ predicted for the $N$ object queries by the preceding decoder layer, we convert the class-class co-occurrence measurements into item-item co-occurrence matrices $R_c\in \mathbb{R}^{N\times N\times3}$ as Eq. \ref{eq6}, where $i$ and $j$ are item indices.
\begin{equation}
R_C = \begin{bmatrix} R_{C_1}, & R_{C_2}, R_{C_3} \end{bmatrix} = \begin{bmatrix}R_{nc}[c_i, c_j], R_{pcc}[c_i, c_j], R_{\tanh(pmi)}[c_i, c_j]\end{bmatrix} 
\label{eq6}
\end{equation}
 
\subsection{Item-Item and Item-Human Relative Position Matrices}\label{subsec3_2}
We construct item-item relative position matrices to capture the relative positional and scale relationships between items. Given the bounding boxes $b_i=[x_i,y_i,w_i,h_i]$ and $b_j=[x_j,y_j,w_j,h_j]$ estimated for two object queries, we represent their spatial relation by the normalized distances and relative scales, as Eq. \ref{eq7}, where $R_P \in \mathbb{R}^{N\times N \times 4}$. 
\begin{equation}
R_P[i,j] = \begin{bmatrix}
\frac{x_i - x_j}{w_i},\frac{y_i - y_j}{h_i},\frac{w_i}{w_j},\frac{h_i}{h_j}
\end{bmatrix}
\label{eq7}
\end{equation}

We also incorporate the relative positional information of each item with respect to human body key-points, which are predicted by the human pose estimator. The pose estimator was trained for the COCO pose dataset, which contains annotations of 17 key-points, including the shoulders, elbows, wrists, hips, knees, and ankles. We refer only to the detection results for some of them to reduce complexity. We selected four types of key points—shoulder, hip, knee, and ankle—because they provide useful spatial references for distinguishing certain item classes. Since these key-points are detected as pairs, we first compute their average coordinates as Eq. \ref{eq8}.
We use only their vertical coordinates, as individuals in outfit images are mostly depicted in standing or sitting postures, with fashion items predominantly aligned vertically.
\begin{equation}
h_k = (x_k, y_k) = \left( \frac{x_{k_L} + x_{k_R}}{2}, \frac{y_{k_L} + y_{k_R}}{2} \right), k \in \{shoulder, hip, knee, ankle\}\
\label{eq8}
\end{equation}

A potential issue is that certain key-points may remain undetected or be detected unreliably, which could negatively affect the detection process. To mitigate this issue, we consider only the key-points with high confidence scores. Letting the average score of the left and right key-points be denoted as $s_k$, the relative distance between the center point of each bounding box and the average key-point is computed as shown in Eq. \ref{eq9}, where, $i$ and $j$ are item indices, while $k$ is the key-point index. We set $\theta = 0.3$.
\begin{equation}
R_{S_k}[i, j] = 
\begin{cases} 
\frac{y_i - y_k}{y_j - y_k}, & \text{if } s_{k} \geq \theta , \\
0, & \text{otherwise}.
\end{cases}
\label{eq9}
\end{equation}

Consequently, the item-human relative position matrix $R_S \in \mathbb{R}^{N\times N \times 4}$ is computed as Eq. \ref{eq10}.
\begin{equation}
R_S = \begin{bmatrix}
R_{S_{shoulder}}, R_{S_{hip}}, R_{S_{knee}}, R_{S_{ankle}}
\end{bmatrix}
\label{eq10}
\end{equation}




\section{Experiments}\label{sec4}

\subsection{Dataset and Experimental Settings}\label{subsec4_1}
For training and evaluation, we utilized the Showniq-H dataset, which is a subset of Showniq, an in-house dataset developed by DeepFashion Co., Ltd., a fashion platform company. The Showniq dataset contains 160K diverse fashion images categorized into 87 item classes. This dataset was collected to support the development of a commercial online fashion platform and provides a realistic representation of fashion items. It includes a broad range of images, encompassing both individual item images and complete outfit images.

\begin{figure}[ht!]
\centering
\includegraphics[width=\columnwidth]{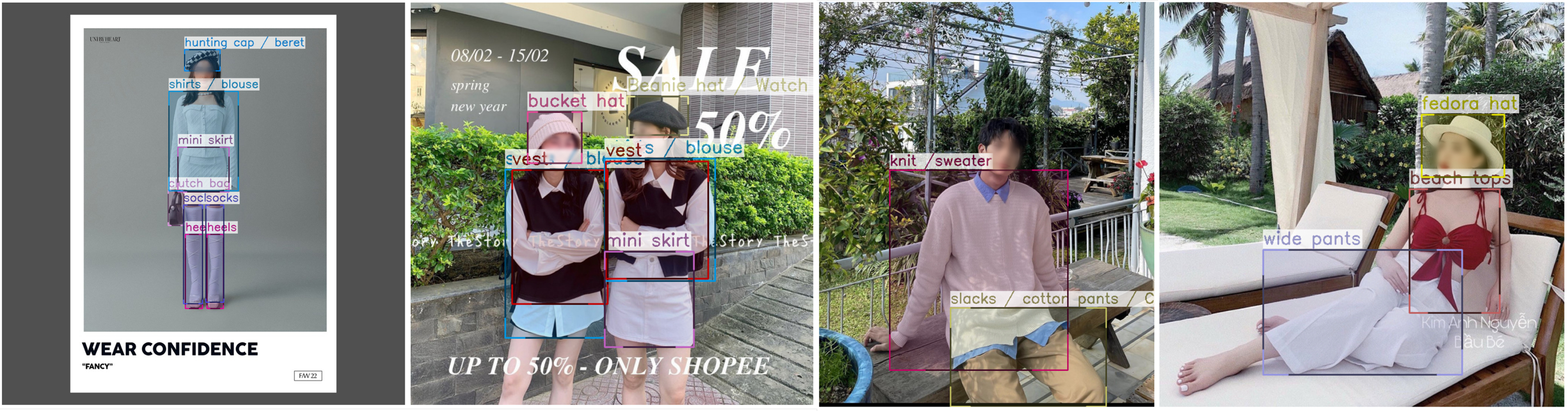}
\caption{Examples of fashion images in the Showniq-H dataset}
\label{fig4}
\end{figure}

Showniq-H is a subset of Showniq, curated specifically for contextual fashion item detection. It consists exclusively of outfit images containing two or more items, with accessory classes excluded. The dataset comprises a total of 67,139 samples, categorized into 12 top-level categories and 80 subcategories, as detailed in Table \ref{tab1}. Showniq-H provides ground-truth bounding boxes and class labels for fashion items. The median image size is $960 \times 1200$. Fig. \ref{fig4} presents several examples of fashion images in Showniq-H. The dataset is divided into 48,411 training samples, 5,397 validation samples, and 13,331 test samples.
    

We developed Holi-DETR based on two baseline models: vanilla DETR \cite{carion2020end} and the more recent Co-DETR \cite{zong2023detrs}. We refer to the variant of Holi-DETR based on Co-DETR as \textit{Holi-Co-DETR}. We implemented based on their official source codes \cite{facebook2020detr,sensex2024codetr} and trained the models following their default settings. For DETR, we trained the model for 300 epochs using the AdamW optimizer with an initial learning rate of $1\times10^{-4}$ (reduced to $1\times10^{-5}$ for the backbone) and a weight decay of $1\times10^{-4}$. Similarly, we trained the Co-DETR models with the AdamW optimizer, employing a weight decay of $1\times10^{-4}$. Gradient clipping was applied with a threshold of 0.1. The training data was augmented using multi-scale cropping and resizing as well as random flipping. We used ResNet50 \cite{he2016deep} as the backbone for DETR and Swin-Small (Swin-S) \cite{liu2021swin} for Co-DETR. In both baselines, the encoder and decoder each comprised 6 layers.

We trained on two NVIDIA GeForce RTX 4090 GPUs in a distributed data parallel (DDP) setup, with a total batch size of 16 (8 samples per GPU) for DETR and with a total batch size of 4 (2 samples per GPU). For pose estimation, we used ViTPose \cite{xu2022vitpose}, which was pretrained on the COCO key-point detection dataset. 

\subsection{Quantitative Evaluation}\label{subsec4_2}

First, we compared the performance of two variants of Holi-DETR models, one based on vanilla DETR and the other on Co-DETR, against their respective baseline models using average precision (AP) metrics. Table~\ref{tab2} presents the experimental results. Holi-DETR demonstrated improved performance over vanilla DETR across all metrics. In particular, Holi-DETR achieved an AP of 56.2\%, which is 3.6 percent points (pp) higher than that of vanilla DETR. Similarly, Holi-Co-DETR achieved an AP of 66.2\%, showing an improvement of 1.1  pp over Co-DETR.
These improvements demonstrate the effectiveness of the proposed methods in enhancing detection performance.


\begin{table}[h!]
    \caption{Experimental results}\label{tab2}
    \begin{tabular}{l|lll}
        \toprule
        Models & AP & AP$_{50}$ & AP$_{75}$ \\
        \midrule
        DETR \cite{carion2020end} & 52.5 & 65.8 & 59.1 \\
        Holi-DETR       & \textbf{56.1} \textcolor{blue}{(+3.6)} & \textbf{67.1} \textcolor{blue}{(+1.3)} & \textbf{62.3} \textcolor{blue}{(+3.2)} \\
        \midrule
        Co-DETR \cite{zong2023detrs} & 65.1 & 73.8 & 70.1 \\
        Holi-Co-DETR & \textbf{66.2} \textcolor{blue}{(+1.1)} & \textbf{74.6} \textcolor{blue}{(+0.8)} & \textbf{71.0}  \textcolor{blue}{(+0.9)}\\
        \bottomrule
    \end{tabular}
\end{table}

\begin{figure}[ht!]
\centering
\includegraphics[width=\columnwidth]{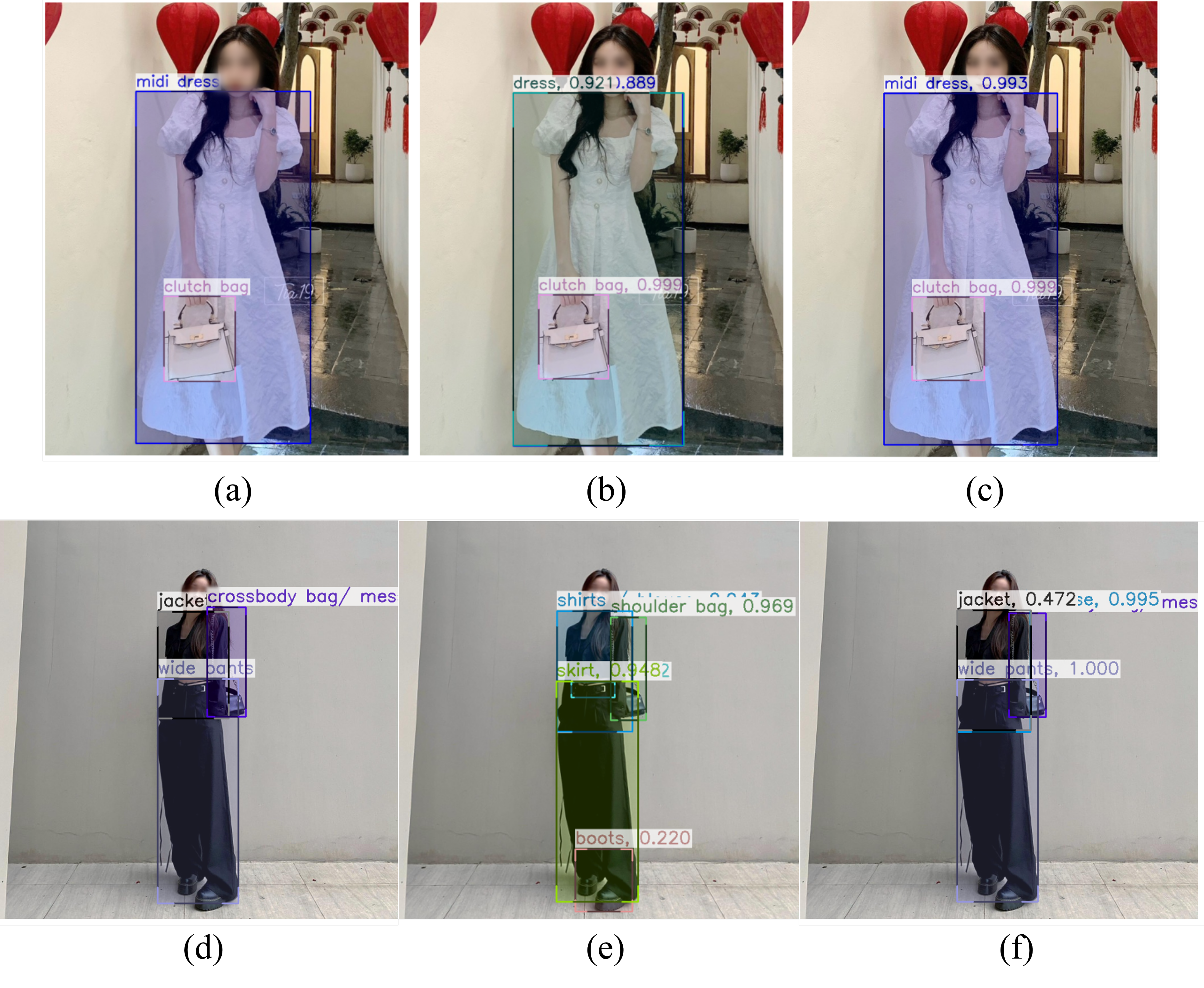}
\caption{
Examples of fashion item detection results. The left, center, and right columns represent the ground-truth labels, DETR results, and Holi-DETR results, respectively
}\label{fig5}
\end{figure}

\begin{figure}[ht!]
\centering
\includegraphics[width=\columnwidth]{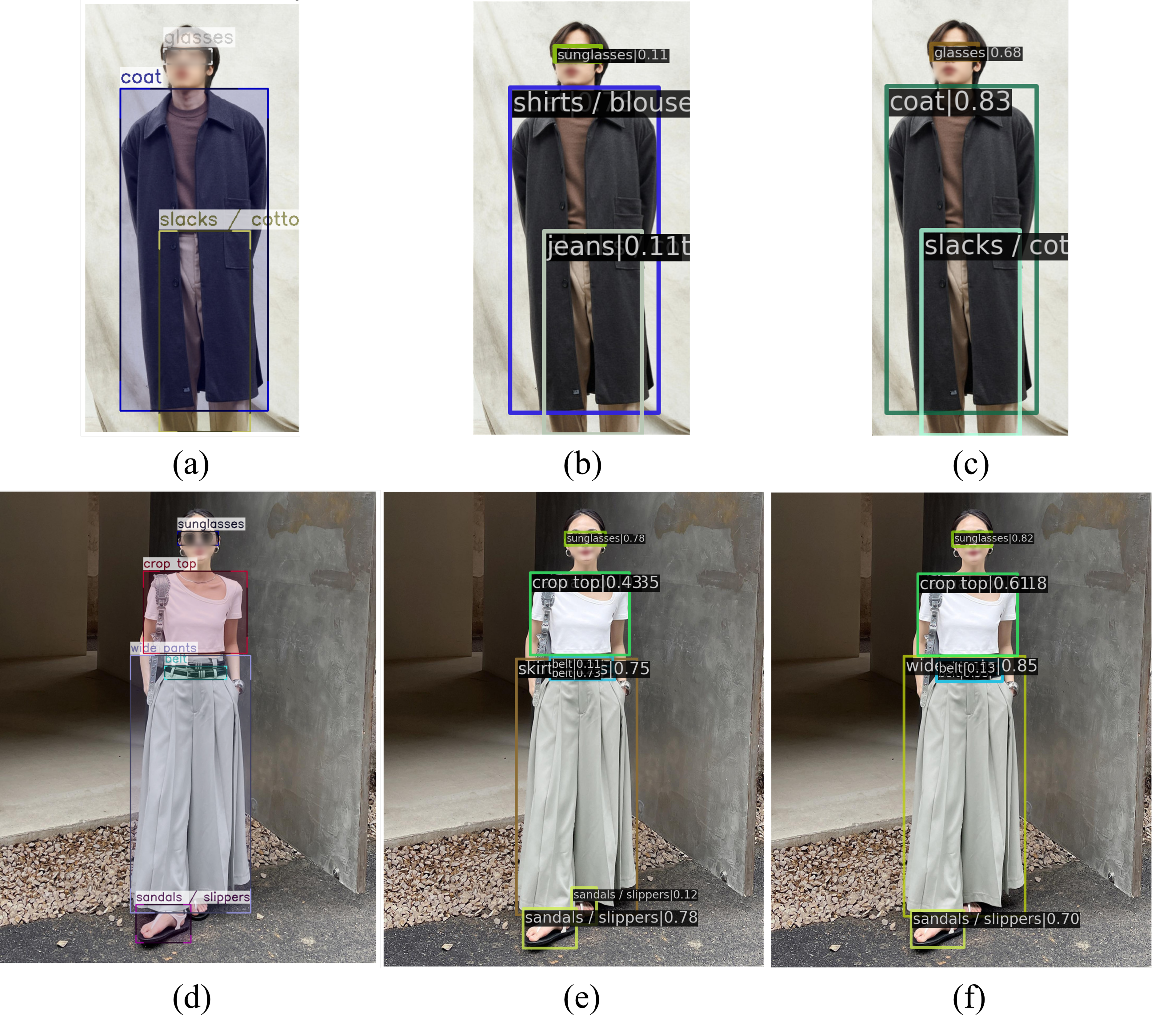}
\caption{Examples of fashion item detection results. The left, center, and right columns represent the ground-truth labels, Co-DETR results, and Holi-Co-DETR results, respectively
}\label{fig6}
\end{figure}

\subsection{Qualitative Evaluation}\label{subsec4_3}
We also manually compared the detection results of the Holi-DETR models with those of their baseline models. Holi-DETR demonstrated improved discrimination ability for subcategory classes that the baseline models struggled to differentiate, as their visual forms were highly similar.
Fig.~\ref{fig5} and ~\ref{fig6} compares the detection results of the models based on vanilla DETR and Co-DETR, respectively.

Fig.~\ref{fig5}(a) presents a person wearing a `midi dress' and a `clutch bag'. In (b), DETR misclassified the `midi dress' as `dress' with a high confidence score of 0.921. DETR also generated another candidate with the class label `dress' and nearly identical coordinates. However, its confidence score was 0.889, slightly lower than that of the `midi dress' candidate. This confusion between the `midi dress' and `dress' classes is likely due to their visual similarity, which can make them indistinguishable in certain outfit images. In contrast, in (c), Holi-DET successfully classified the item as a `midi dress' with a high confidence score of 0.993 by leveraging the relative position and size of each item with respect to other items or human body key points, such as the knee and ankle.

In Fig.~\ref{fig5}(e), DETR misclassified the `wide pants' as `skirt' and the `crossbody bag/messenger bag' as `shoulder bag', respectively. It is presumed that the `wide pants' were misclassified due to their dark color, the illumination conditions, and the wearer's pose, which obscured the boundary between the two legs. Similarly, the `crossbody bag/messenger bag' is visually very similar to a `shoulder bag,' making it prone to misclassification. However, in Fig.~\ref{fig5}(f), Holi-DETR successfully recognized the `wide pants' and `crossbody bag/messenger bag' with high confidence scores of 0.997 and 1.0, respectively. Regarding the `jacket' in (d), both models misclassified it. Nevertheless, while DETR failed to generate any candidate labeled `jacket', Holi-DETR produced a candidate of `jacket' with a confidence score of 0.472.


In Fig.~\ref{fig6}(a), the man is wearing a `coat', `glasses', and `slacks/cotton pants/chino'. In (b), Co-DETR misclassified the `coat' as `shirts/blouse', likely due to the similarity of the `coat' silhouette to that of shirts. However, in (c), Holi-Co-DETR accurately classified it as a `coat'. Similarly, the `glasses' in (a) was misclassified as `sunglasses' by Co-DETR, but Holi-Co-DETR correctly identified it.

The woman in Fig.~\ref{fig6}(d) is wearing a `crop top', However, Co-DETR barely distinguished it from `casual-top', assigning both relatively low confidence scores of 0.430 and 0.353, respectively. Holi-Co-DETR, on the other hand, provided a significantly higher confidence score of 0.614 for the correct `crop top' label. Furthermore, in (e), whereas Holi-Co-DETR accurately identified `wide pants', Co-DETR confused the same item as a `skirt'. As demonstrated in Fig.~\ref{fig5} and ~\ref{fig6}, Holi-DETR models exhibits improved discrimination capability between classes that share relatively similar appearances.


\subsection{Ablation studies}\label{subsec4_4}

\begin{table}[h!]
    \caption{The results of ablation studies}\label{tab3}
    \begin{tabular}{l|ccc|ccc}
        \toprule
        Models & Co-occurrence & Item-Item Rel Pos & Item-Human Rel Pos & AP & AP$_{50}$ & AP$_{75}$ \\
        \midrule
        \multirow{6}{*}{DETR \cite{carion2020end}} & & & & 52.5 & 65.8 & 59.1 \\
        & \checkmark & & & 55.8 & 66.7 & 61.6 \\
        & \checkmark & & \checkmark & \textbf{56.1} & \textbf{67.1} & \textbf{62.3} \\
        \cmidrule(lr){2-7}
        & & \checkmark & & 55.6 & 66.3 & 61.3 \\
        & \checkmark & \checkmark & & 55.9 & 66.9 & 62.0 \\
        & \checkmark & \checkmark & \checkmark & 55.9 & 67.0 & 61.8 \\
        \midrule
        \multirow{6}{*}{Co-DETR \cite{zong2023detrs}} & & & & 65.1 & 73.8 & 70.1 \\
        & \checkmark & & & 65.9 & 74.4 & 70.6 \\
        & \checkmark & & \checkmark & \textbf{66.2} & \textbf{74.6} & \textbf{71.0} \\
        \cmidrule(lr){2-7}
        & & \checkmark & & 65.5 & 74.0 & 70.1 \\
        & \checkmark & \checkmark & \checkmark & 66.1 & 74.5 & 70.9 \\
        \bottomrule
    \end{tabular}
\end{table}

We conducted ablation studies to analyze the contribution of the three types of contextual information. We selectively enabled or disabled the three branches of the context encoder in Fig.~\ref{fig3}, each of which is to encode one of the three types of contextual information.

When incorporating item-item co-occurrence relationships into DETR, the AP increased from 52.5\% to 55.8\%, reflecting an improvement of 3.3 pp. Additionally, referencing item-human relative position further improved the AP by 0.3 pp. When item-item relative position was referenced alone, the AP improved from 52.5\% to 55.6\%, representing a 3.1 pp gain compared to the baseline model. On the other hand, when item-item relative position was used in conjunction with item-item co-occurrence, only a marginal improvement of 0.1 pp was observed, increasing the AP from 55.8\% to 55.9\%. Interestingly, when all three contextual information types were used together, the AP decreased slightly from 56.1\% to 55.9\% by 0.2\%.
This indicates that while item-item relative position contributes to a meaningful improvement, its effect may be redundant when combined with the other two types of contextual information.
The same trend was consistently observed in the ablation studies based on Co-DETR. Consequently, our best AP is 66.2\% achieved by Holi-Co-DETR, which references both co-occurrence relationships and item-human relative position.


\section{Conclusion}\label{sec5}
In this study, we propose a novel approach for holistic fashion item detection that leverages contextual information.  
The proposed method improves detection performance by leveraging the co-occurrence relationships among items and their relative positional information to other items or human body key-points. When applied to vanilla DETR and the more recent Co-DETR, the proposed methods achieved absolute improvements of 3.6\% and 1.1\%, respectively, on the highly diverse Showniq-H dataset.
One limitation of the proposed method is that it relies on co-occurrence counts extracted from the training data. As a result, if the combination of items in the input image is uncommon, it may potentially lead to unintended side effects.
Our work provides a meaningful reference for researchers exploring AI applications in the fashion domain by introducing the first Transformer-based contextual object detection model specifically designed for fashion data.



\bibliography{sn-bibliography}

\end{document}